\documentclass[sigconf]{acmart}

\AtBeginDocument{%
  \providecommand\BibTeX{{%
    \normalfont B\kern-0.5em{\scshape i\kern-0.25em b}\kern-0.8em\TeX}}}

\copyrightyear{2022}
\acmYear{2022}
\setcopyright{acmcopyright}
\acmConference[SIGIR '22]{Proceedings of the 45th International ACM SIGIR Conference on Research and Development in Information Retrieval}{July 11--15, 2022}{Madrid, Spain.}
\acmBooktitle{Proceedings of the 45th International ACM SIGIR Conference on Research and Development in Information Retrieval (SIGIR '22), July 11--15, 2022, Madrid, Spain}
\acmPrice{15.00}
\acmISBN{978-1-4503-8732-3/22/07}
\acmDOI{10.1145/3477495.3531804}
\begin{document}
\fancyhead{}

\title{BSAL: A Framework of Bi-component Structure and Attribute Learning for Link Prediction}


\author{Bisheng Li}
\authornote{Both authors contributed equally.}
\authornote{Work partially done during the internship in Huawei Noah's Ark Lab.}
\affiliation{
  \institution{Fudan University}
  \city{Shanghai}
  \country{China}}
\email{bsli20@fudan.edu.cn}

\author{Min Zhou}
\authornotemark[1]
\authornote{Corresponding Authors.}
\affiliation{
  \institution{Huawei Noah's Ark Lab}
  \city{Shenzhen}
  \country{China}
}
\email{zhoumin27@huawei.com}

\author{Shengzhong Zhang}
\affiliation{
 \institution{Fudan University}
 \city{Shanghai}
 \country{China}
}
\email{szzhang17@fudan.edu.cn}

\author{Menglin Yang}
\affiliation{
 \institution{The Chinese University of Hong Kong}
 \city{Hong Kong}
 \country{China}
}
\email{mlyang@cse.cuhk.edu.hk}

\author{Defu Lian}
\affiliation{
 \institution{University of Science and Technology of China}
 \city{Hefei}
 \country{China}
}
\email{liandefu@ustc.edu.cn}

\author{Zengfeng Huang}
\authornotemark[3]
\affiliation{
 \institution{Fudan University}
 \city{Shanghai}
 \country{China}
}
\email{huangzf@fudan.edu.cn}

\renewcommand{\shortauthors}{Li and Zhou, et al.}

\begin{abstract}
Given the ubiquitous existence of graph-structured data, learning the representations of nodes for the downstream tasks ranging from node classification, link prediction to graph classification is of crucial importance. Regarding missing link inference of diverse networks, we revisit the link prediction techniques and identify the importance of both the structural and attribute information. However, the available techniques either heavily count on the network topology which is spurious in practice, or cannot integrate graph topology and features properly. To bridge the gap, we propose a bicomponent structural and attribute learning framework (BSAL) that is designed to adaptively leverage information from topology and feature spaces. Specifically, BSAL constructs a semantic topology via the node attributes and then gets the embeddings regarding the semantic view, which provides a flexible and easy-to-implement solution to adaptively incorporate the information carried by the node attributes. Then the semantic embedding together with topology embedding are fused together using attention mechanism for the final prediction. Extensive experiments show the superior performance of our proposal and it significantly outperforms  baselines 
on diverse research benchmarks.
\end{abstract}


\maketitle

\section{Introduction}

\begin{figure}[!tp]
    \centering
    \includegraphics[width=0.9\linewidth]{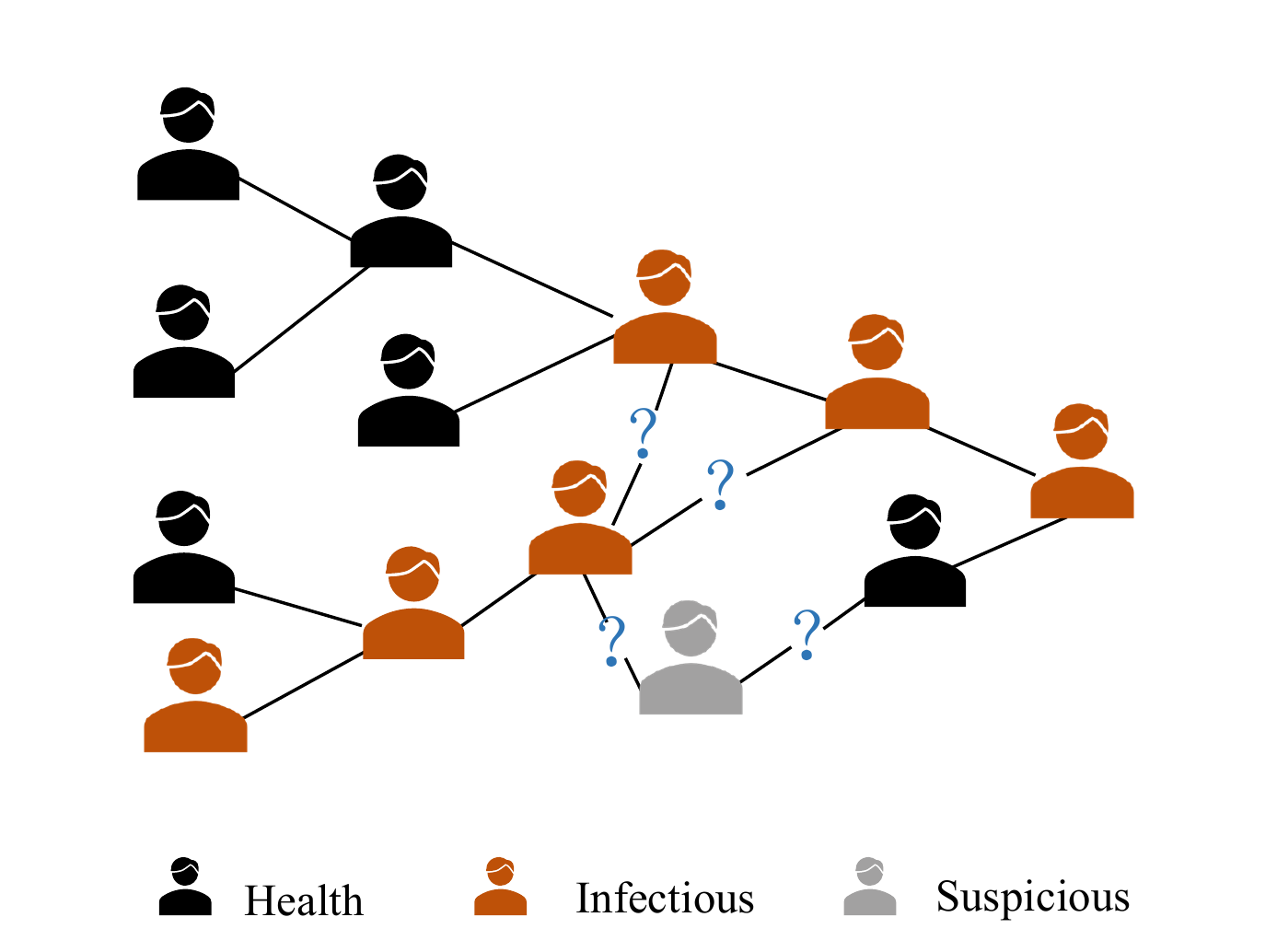}
    \caption{An illustration of disease spreading networks, which is tree-like with a distinct hierarchical layout ~\cite{weiss2013sir}. Inferring the missing links helps to track the disease propagation path or possible interactions during an epidemic break (e.g., COVID-19), providing guidance for the control action or implementation. }
    \label{disease}
    \vspace{-15pt}
\end{figure}

Link prediction aims to predict whether two nodes in a network are likely to have a link \cite{liben2007link,yang2021discrete, wang2021gognn} via the partially available topology or/and node attributes, attracting considerable research efforts owning to its diverse applications, such as the friend recommendation in social networks~\cite{adamic2003friends,wang2014friendbook}, the item recommendation \cite{koren2009matrix,wang2021hypersorec,yang2022hrcf} in user-item networks, location recommendation in urban computing~\cite{lian2020geography,lian2014geomf}, the relation completion in knowledge graphs~\cite{zhang2021cone,Lu2022DensE}, drug-target prediction in biomedical networks~\cite{liu2016neighborhood, ou2021matrix,zhang2021motif}, and disease infectious analysis of epidemic spread~\cite{chami2019hyperbolic,gang2005epidemic,yang2021discrete}.

According to the techniques involved, the current link prediction methods can be categorized into three classes: heuristic methods, embedding methods, and GNN-based methods. Heuristic methods infer the likelihood of links via handcrafted similarity measures regarding the structure information ~\cite{liben2007link,barabasi1999emergence}. Embedding methods, learn free-parameter node embeddings in a transductive manner, thus do not generalize to unseen nodes and networks~\cite{node2vec-kdd2016,koren2009matrix}. Graph neural network (GNN) based methods formulate link prediction as the binary classification problem in which the  explicit node feature could be incorporated~\cite{kipf2016variational,zhang2018link,yun2021neo}.

Two main classes of GNN-based link prediction methods are Graph AutoEncoder (GAE)~\cite{kipf2016variational} and SEAL~\cite{zhang2018link,zhang2021labeling}. However,  it's empirically observed that GAE which heavily relies on smoothed node features shows poor performance on the datasets with highly hierarchical layouts, which limits its application in many real-world scenarios with a distinct tree-like structure~\cite{chami2019hyperbolic,yang2022hyperbolic,liu2022enhancing} such as disease spreading networks (as sketched in Fig. \ref{disease}).
On the other hand, SEAL applies a specific labeling trick to explicitly encode the structural information of the enclosing subgraphs around each link and then applies a graph-level GNN, demonstrating a more powerful link inference ability. SEAL provides a promising fashion for link predictions, but the pioneering work heavily counts on the network topology which is inherently incomplete and spurious under the link prediction settings. Though the node attributes are suggested into the joint learning process of SEAL, directly combining the node features with the node structural matrix does not always improve the performance as they are from two different domains which are distant in both information formats as well the dimensionality.

To alleviate the limitations of GAE and SEAL, we propose a bi-component structure and attributes learning framework (BSAL) that is designed to adaptively integrate key information from both topology and feature domains. Specifically, BSAL constructs a semantic topology via the node features and then gets the structural embeddings regarding the semantic view, which provides a flexible and easy-to-implement solution to adaptively incorporate the node attributes. Then BSAL measures the existence of links by considering fusing the information of the original topology as well as the semantic topology.  We show the effectiveness and superiority of BSAL as it significantly and consistently outperforms diverse baselines on various research benchmarks.

\section{Methodology}
\label{methodology}

By following the subgraph classification paradigm, the current SOTA methods, SEAL and LGLP obtain great success in link prediction tasks. However, the pioneering works~\cite{zhang2018link,zhang2021labeling,cai2021line} solely utilize local topology information and fail to leash the rich semantic information carried by the nodes attributes, which hinders both the performance improvement and the industrial applications. 

\begin{figure*}[!tp]
    \centering
    \includegraphics[width=0.99\linewidth]{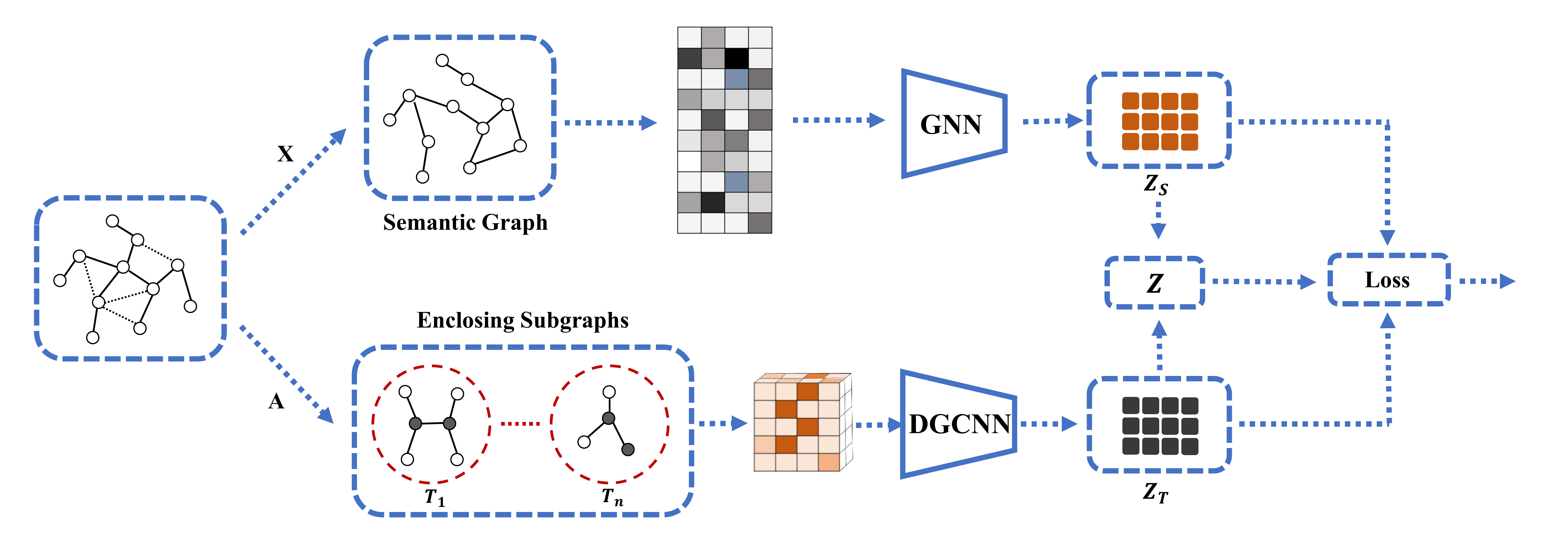}
    \caption{The framework of the BSAL model which contains two channels. Adjacency  matrix A is used to extract $n$ subgraphs and  learn the corresponding feature matrices via the node distance labeling.  Node attribute X is utilized to construct a semantic graph and then provides auxiliary structural information to assist the subgraph-based link inference.}
    \label{framework}
    \vspace{-10pt}
\end{figure*}

To alleviate the limitations, we propose a bi-component link prediction framework, named BSAL, that provides a flexible and efficient solution for incorporating the informative and bountiful graph topology and node attributes. As structural information is crucial to link prediction tasks~\cite{zhang2018link,cai2021line}, our BSAL takes the powerful subgraph classification schema(i.e. SEAL, LGLP) as the backbone. Priority of SEAL mainly lies in the extraction of local structural information and the key idea of BSAL is to construct semantic topology from the node attributes.  It translates the semantic information to the global-level structural features to compensate for the shortcomings. The overall framework is illustrated in Figure~\ref{framework}. 

\medskip
Giving a graph with node attribute $X$ and adjacency matrix $A$. First, we construct semantic topology $G_s$ with $X$ and then extract the semantic features $z_{s,i}$ of each node $i$. After that, the semantic features along with the structural (topological) features of each subgraph are finetuned using two different GNNs. The weighted sum of embeddings from the topology space and the feature space are used to learn the probability of link existence of the target node pair. The key steps are illustrated as follows.

\begin{table}[t]
\centering
\caption{A Summary of Similarity Measures.}
\resizebox{0.3\textwidth}{!}{%
\small
\begin{tabular}{@{}l|l@{}}
\toprule
\textbf{Name}        & Score function \\ \midrule \midrule
\textbf{Inner Product} & $   S_{ij} = x_{i} \cdot x_{j}$      \\
\textbf{Euclidean Distance} &
$S_{ij} = - \| x_{i} - x_{j} \|^{2} $ \\
\textbf{Cosine Similarity}&  $S_{ij} = \frac{x_{i} \cdot x_{j}}{\|x_{i}\| \cdot \|x_{j}\|}$   \\
\textbf{Gaussian Kernel}  & $S_{ij} = \exp ( - \frac{\|x_{i} - x_{j}\|}{t})$ \\   
\bottomrule
\end{tabular}%
}
\label{tab:similarity}
\vspace{-5pt}
\end{table}

\medskip
\textit{\textbf{Semantic Topology Construction. }}
To facilitate subgraph-based structural learning via the node attributes, we suggest constructing a new semantic graph $G_{s}$, by which the attributes information could be easily translated to the structural level embeddings. What's more, the node attributes are diverse formats in many real applications. The way that builds a semantic graph is easily adapted to the applications regarding the domain knowledge.

Given datasets(e.g. research benchmarks) where the node attributes are feature vectors, the semantic graph can be got by directly calculating the similarity of the features of given pair nodes via a proper measure and appointing the $k$ nodes with higher scores of a target node as its neighbors~\cite{wang2020gcn,wu2020connecting}  \footnote{$k$ is the hyperparameter and the average degree of each dataset is the default in the experiments for simplicity.}. Several commonly used scoring functions are summarized in Table \ref{tab:similarity}, in which $x_{i}$ and $x_{j}$ represent the feature vector of node $i$ and $j$. Specifically, Euclidean Distance is employed in this work.

\medskip
\textit{\textbf{Semantic Feature Extraction. }}
After obtaining the semantic topology $G_{s}$, we aim to extract structural information from the semantic graph to further combine with the distance encoding matrix $Z_{T,j}$ of a given subgraph $j$. As the semantic graph is intrinsically constructed by the feature matrix, structure-based graph embeddings techniques which give the node embeddings regarding the topology are suitable~\footnote{ In this work, the 
node2vec \cite{node2vec-kdd2016} which learns feature representations via optimizing a neighborhood preserving objective is employed.}.To this end, structural embedding matrix $Z_S$ regarding the global structural information of the semantic graph is obtained with the embedding of node $i$ denoted by  $z_{s,i}$. 

\medskip
\textit{\textbf{Feature Finetuning. }}
After distance encoding of subgraphs and semantic feature extraction, we have embeddings from different topology and feature spaces. Directly concatenating them together may hinder the model performance since they are from different domains. So before we integrate these coarse-grained features, we finetune the embeddings using two different graph neural networks, respectively. Then, we transform topological embeddings $Z_{T}$ and semantic embedding $Z_{S}$ into fine-grained embedding and the downstream task can utilize them directly without considering their source domains.

\medskip
\textit{\textbf{Training. }}
Based on the topological embedding $Z_{T}$ and semantic embedding $Z_{S}$, we focused on the node pair $(i, j)$, where its embeddings are $Z_{(i, j)}^{T} \in \mathbb{R}^{h\times1}$ and $Z_{(i, j)}^{S} \in \mathbb{R}^{h\times1}$, we first transform the embedding through a linear transformation $W \in \mathbb{R}^{h^{'}\times h}$ and follows up with a nonlinear activation function, and then use one shared attention head $q \in \mathbb{R}^{h^{'} \times 1}$ to get the attention value $\omega_{(i, j)}$ as follows:
\begin{equation}
    \omega_{(i, j)} = q^{T} \cdot \tanh(W \cdot Z_{(i, j)} + b).
\end{equation}
We then normalize the attention values using the softmax function to get the final weight $\alpha_{(i, j)}$. The final embedding for node pair $(i, j)$ can be formulated as:
\begin{equation}
    Z_{(i, j)} = \alpha_{(i, j)}^{T} \cdot Z_{(i, j)}^{T} + \alpha_{(i, j)}^{S} \cdot Z_{(i, j)}^{S}.
\end{equation}
Based on (\ref{lossf}), we jointly train our proposal using three standard binary cross entropy losses:
\begin{equation}
    \mathcal{L} = \sum_{(i, j) \in D} \alpha \cdot BCE(Z_{(i, j)}^{T}, y_{ij}) + \beta \cdot BCE(Z_{(i, j)}^{S}, y_{ij}) + BCE(Z_{(i, j)}, y_{ij}),
    \label{lossf}
\end{equation}
where $y_{ij}$ denotes the link existence and $BCE(\cdot, \cdot)$ is the binary cross entropy loss, $\alpha$ and $\beta$ are hyperparameters that weigh the importance of the corresponding loss terms. \footnote{The default values of  $\alpha$ and $\beta$ are 1}.

\begin{table}[h]
\centering
\caption{Statistics of the datasets.}
\resizebox{0.43\textwidth}{!}{%
\begin{tabular}{@{}lcccc@{}}
\toprule
{\sc Dataset}  & Nodes & Edges & Classes & Node features \\ \midrule
{\sc Disease}  & 1044  & 2663  & 2       & 1000          \\
{\sc Citeseer} & 4230  & 10674  & 6       & 602          \\
{\sc Cora}     & 2708  & 10556  & 7       & 1433          \\

{\sc Twitch\_en}   & 7126 & 77774 & 2       & 128           \\
{\sc CoauthorCS} & 18333 &163788 & 15 & 6805             \\
\bottomrule
\end{tabular} 
}
\label{tab:statistics}
\vspace{-5pt}
\end{table}

\section{Experiments}
\label{experiments}

\begin{table*}[htp]
    \centering
    \caption{Comparison with baseline models. All the experiments have been conducted for 10 times and we take the average value as results, and the standard deviations are also listed. For both the two metrics, the higher, the better. The best are in bold and the second best are underlined.}
    \resizebox{1.0\textwidth}{!}{%
    \begin{tabular}{@{}lccccc|ccccc@{}}
    \toprule
     & \multicolumn{5}{c|}{AUC} & \multicolumn{5}{c}{AP} \\ \midrule
    Model & {\sc Disease} & {\sc CiteSeer} & {\sc Cora} & {\sc Twitch\_EN} & {\sc CoauthorCS} & {\sc Disease} & {\sc CiteSeer} & {\sc Cora} & {\sc Twitch\_EN} & {\sc CoauthorCS} \\ \midrule
    CN & 50.00 & 63.23 & 71.66 & 76.29 & 89.56 & 50.00 & 63.18 & 71.33 & 75.99 & 89.51 \\
    AA & 50.00 & 63.25 & 71.68 & 76.57 & 89.59 & 50.00 & 63.30 & 71.55 & 77.17 & 89.63 \\
    PPR & 37.59 & 71.63 & 82.86 & 84.14 & 95.55 & 50.00 & 74.79 & 88.12 & 87.40 & 97.12 \\ \midrule
    \textit{node2vec} & 44.64 $\pm$ 0.94  &  74.62  $\pm$ 1.00  &  86.02  $\pm$ 0.68  &  86.43  $\pm$ 0.12  &  96.26  $\pm$ 0.05 & 52.07 $\pm$ 1.13 & 82.05 $\pm$ 0.57 & 89.01 $\pm$ 0.40 & 89.81 $\pm$ 0.10 & 96.92 $\pm$ 0.03 \\
    GCN &  75.93  $\pm$ 1.95  &  90.86  $\pm$ 0.45  &  88.52  $\pm$ 1.06 &  86.20  $\pm$ 0.24  &  94.73  $\pm$ 0.24 & 78.60 $\pm$ 1.45 & 91.47 $\pm$ 0.43 & 88.37 $\pm$ 1.37 & 88.61 $\pm$ 0.80 & 94.56 $\pm$ 0.24 \\
    GAT &  70.65  $\pm$ 1.19  &  89.60  $\pm$ 0.52  &  89.25  $\pm$ 0.77   &  82.30  $\pm$ 0.26  &  95.28  $\pm$ 0.08 & 70.99 $\pm$ 1.70 & 89.31 $\pm$ 0.67 & 89.17 $\pm$ 0.87 & 83.64 $\pm$ 0.47 & 94.91 $\pm$ 0.14 \\ \midrule
    SEAL(wo feat) &  95.23 $\pm$ 0.05  &  91.62  $\pm$ 0.15  &  \underline{91.01}  $\pm$ \underline{0.14}  & 91.01  $\pm$ 0.04  &  96.51  $\pm$ 0.04 & 91.94 $\pm$ 0.05 & 92.46 $\pm$ 0.16 & 91.95 $\pm$ 0.11 & 92.07 $\pm$ 0.06 & 97.44 $\pm$ 0.03 \\
    SEAL(w feat) &  \underline{96.88}  $\pm$ \underline{0.77}  &  89.86  $\pm$ 0.77  &  88.33  $\pm$ 0.78  & \underline{92.15} $\pm$ \underline{0.18} & 96.64 $\pm$ 0.27 & \underline{95.35} $\pm$ \underline{1.14} & 91.35 $\pm$ 0.66 & 89.58 $\pm$ 0.62 & \underline{92.97} $\pm$ \underline{0.24} & 97.28 $\pm$ 0.22  \\ 
    LGLP &  87.52  $\pm$ 0.00  &  \underline{92.52}  $\pm$ \underline{0.07}  &  90.86  $\pm$ 0.33  &  91.01  $\pm$ 0.05  &  96.06  $\pm$ 0.00 & 80.06 $\pm$ 0.00 & \underline{93.19} $\pm$ \underline{0.08} & \underline{91.72} $\pm$ \underline{0.32} & 92.43 $\pm$ 0.03 & 97.13 $\pm$ 0.01\\ 
    NeoGNN &  82.13  $\pm$ 1.73  &  91.22  $\pm$ 0.64  &  90.96  $\pm$ 0.86  &  90.08  $\pm$ 0.67  &  \textbf{97.90}  $\pm$ \textbf{0.19} & 82.69 $\pm$ 1.35 & 92.86 $\pm$ 0.55 & 91.58 $\pm$ 0.39 & 91.71 $\pm$ 0.68 & \textbf{98.23} $\pm$ \textbf{0.07} \\ \midrule
    BSAL(ours) &  \textbf{97.39} $\pm$ \textbf{0.00} & \textbf{93.24} $\pm$ \textbf{0.05} & \textbf{91.11} $\pm$ \textbf{0.10} & \textbf{92.81} $\pm$ \textbf{0.05} & \underline{96.79} $\pm$ \underline{0.06} &  \textbf{97.21} $\pm$ \textbf{0.01} & \textbf{94.11} $\pm$ \textbf{0.07} & \textbf{92.21} $\pm$ \textbf{0.1} & \textbf{93.31} $\pm$ \textbf{0.05} & \underline{97.61} $\pm$ \underline{0.06}\\\bottomrule
    \end{tabular}
    }
    \label{tab:my_label}
    \vspace{-5pt}
\end{table*}

\subsection{Experimental Setup}

\textbf{Datasets. } To verify the effectiveness , we apply our proposal to diverse research benchmark datasets, whose statistics are summarized in the Table~\ref{tab:statistics}. The dataset sources are also supplemented for reproducibility. The research benchmark datasets used for evaluation in this work are {\sc Disease}~\cite{chami2019hyperbolic}, {\sc Twitch\_en}~\cite{rozemberczki2021multi}, {\sc CoauthorCS}~\cite{shchur2018pitfalls} and two benchmark citation networks, {\sc Citeseer}~\cite{yang2016revisiting} and {\sc Cora}~\cite{huang2021scaling}. In {\sc Disease}, the label of a node indicates whether the node is infected by a disease or not, and the node features associate the susceptibility to the disease. The disease spreading network has an evident hierarchical structure. In {\sc Twitch\_en}, nodes represent gamers on Twitch and edges are fellowships between them. Node features represent embeddings of games played by the Twitch users. {\sc CoauthorCS} is a coauthor network, where nodes represent authors that are connected by an edge if they co-authored a paper. For {\sc Citeseer}, {\sc Cora}, each node is a scientific paper characterized by the corresponding bag-of-words representations, edges denote citations and node labels are academic (sub)areas. Compared to {\sc Disease}, the citation networks are less hierarchical, which are used here to then demonstrate the generalization ability of the proposal.

\textbf{Baselines. } The proposal is compared against the following methods, including heuristic baselines: common neighbors (CN)~\cite{liben2007link}, Adamic-Adar (AA)~\cite{liben2007link}, Personalized PageRank (PPR)~\cite{wang2020personalized}; embedding-based models: \textit{node2vec}; GAE-like models: GCN and GAT; and the state-of-the-art models: SEAL, LGLP, and NeoGNN. Heuristic methods formulate specific rules as scoring functions. \textit{node2vec}~\cite{node2vec-kdd2016} learns representations on graphs via optimizing a neighborhood preserving objective. For GAE-like baselines, we use different encoders including GCN~\cite{kipf2017semi} and GAT \cite{velickovic2018graph}. Their differences are mainly in the mechanism of message passing. GCN aggregates information through structure information while GAT aggregates information through feature correlation. By comparing them, we can show that structure and features are both essential for link prediction. SEAL~\cite{zhang2021labeling}, LGLP~\cite{cai2021line}, and NeoGNN~\cite{yun2021neo} have achieved various SOTA results on link prediction. The priority of SEAL and LGLP mainly lies in the extraction of local structural information while NeoGNN leverages information from both features and topology. Compared with them, it shows the importance of semantic information and the excellence of the proposed method.

\textbf{Hyperparameter settings. } We follow the standard link prediction split ratio and split the edges to 85\%/5\%/10\% for training, validation, and test, respectively. In order to make the comparison fair, we fix the random seed as 2 when splitting the dataset.  The batch size is 32 for all the datasets. We adopt the early stopping strategy with the training epochs as 400, and the patience is set as 20. Each experiment is conducted 10 times and we take the average value as the final result. 
All the code is implemented using PyTorch and we use the implementation in PyTorch Geometric \cite{Fey/Lenssen/2019} for all the models.

\textbf{Evaluation Metrics. } Two standard metrics Area Under Curve (AUC) ~\cite{zhou2009predicting} and Average Precision (AP) ~\cite{Zhang2009} are adopted to evaluate the performance of link prediction methods or algorithms.

\subsection{Experiment Results}
We verify the generalization ability of our proposal via diverse research benchmark datasets, the results are summarized in Table \ref{tab:my_label}. The code and data are available at \url{https://github.com/BishengLi0327/BSAL}. Note that GCN and GAT are all under GAE settings and we use the \textit{InnerProduct} as a decoder to compute the link existence, SEAL (wo feat) denotes the subgraph features do not involve node features, while SEAL (w feat) implies that we directly concatenate node features with structural encoding features together. 

According to the experimental results, we have the following observations. First, the heuristic methods (i.e., CN, AA, and PPR) and graph embedding (i.e., Node2Vec) performs much poorly in general, which confirms the superiority of the GNN-based learning paradigm. Second, SEAL, LGLP, NeoGNN, and BSAL show apparent advantages over the others, demonstrating the importance of structural information in the link prediction task.  
It's also noted that the message-passing methods (i.e., GCN and GAT) all perform poorly on the highly hierarchical dataset (i.e., {\sc Disease}). 
Though NeoGNN alleviates the problem by learning structural information via the adjacency matrix, its performance gap to the subgraph-based methods(e.g, SEAL and BSAL) is still significant as the GAE branch may hinder the expression power of the model.  It is further observed  that SEAL(w feat) shows considerable performance degradation over SEAL(wo feat) on a few datasets (i.e., {\sc Citeseer} and i.e.,{\sc Cora}). In other words, directly combining node features with the position encoding doesn't always give performance gain and sometimes even hinder the model performance in the powerful subgraph-based method. Better ways to incorporate feature information are needed.While the proposed BSAL achieves four best performance overall datasets, it confirms our proposal successfully utilizes the semantic information and paves the way for further research in this field. 

In summary, GNN-based solutions which formulate the link prediction task as a supervised learning problem and learn the node representations via the information aggregation are promising.  The structural information is crucial but not sufficient in the missing link inference. Explicitly encoding the semantic information further contributes to the performance gain.

\section{Related Work}

Existing work on link prediction can be generally classified into three main paradigms: heuristic methods, embedding methods, and GNN-based methods. Heuristic methods compute some heuristic node similarity scores as the likelihood of links~\cite{liben2007link, lu2011link, yang2020featurenorm}, which are simple yet effective. Typical heuristic methods include common neighbors (CN)~\cite{liben2007link}, Adamic-Adar (AA)~\cite{liben2007link} and Personalized PageRank (PPR)~\cite{wang2020personalized}. Although working well in practice, heuristic methods are designed based on handcrafted metrics and may not be applicable for diverse scenarios. Embedding methods learn node embeddings based on connections between nodes and compute similarity scores using the embeddings. LINE ~\cite{tang2015line} learns node embeddings by preserving first and second-order proximity. Random walk-based embedding methods such as DeepWalk~\cite{perozzi2014deepwalk} and \textit{node2vec}~\cite{node2vec-kdd2016} learn node embeddings by applying the Skip-Gram~\cite{church2017word2vec} techniques on the random walks. Recently, considerable efforts have been conducted to apply GNN to link prediction tasks. GAE and VGAE~\cite{kipf2016variational} learn node representations through an auto-encoder framework where various GNN architectures have been utilized as the encoders. NeoGNN ~\cite{yun2021neo}  recognizes the importance of structures in link prediction and considers information from both feature space and topology space, separately. More specifically,  it learns the structural embeddings from the adjacency matrix by a GNN, which are combined with the node representation produced by the  GAE for further prediction. SEAL~\cite{zhang2018link,zhang2021labeling} and LGLP~\cite{cai2021line} reformulated the link prediction task as the binary classification of subgraphs. Instead of directly predicting the existence of a link, SEAL extracts enclose subgraphs around each target link to compose the dataset and perform the graph classification task via a graph-level GNN. While LGLP follows the same subgraph paradigm and converts the link prediction problem into a node classification problem in its corresponding line graph. As the structural information has been explicitly encoded, SEAL and LGLP show apparent advantages over GAE in most cases. 

\section{Conclusion}
In this work, we investigate the link prediction problem in complex network analysis. By revisiting the available techniques, we find the current SOTA GNN-based methods either heavily count on the network topology or node attributes, which is insufficient for highly hierarchical or sparse datasets. To mitigate the problem, we propose a bicomponent structural and attribute learning framework (BSAL) that is designed to fuse key information regarding both topology and attribute spaces.   
Specifically, BSAL constructs a semantic topology via the node attributes and then gets the semantic embeddings regarding the new topology, which provides a flexible and easy-to-implement solution to adaptively incorporate both the topology and the features. It significantly outperforms competing baselines on diverse research benchmarks, which confirms the effectiveness of our proposal.
In the future, we will test the proposal on more datasets, evaluation metrics (e.g., Hits@K), or tasks (e.g., adversarial attacks) to validate its generalization ability. Learning the semantic and topology embeddings in an end-to-end fashion is also a direction to be explored.

\section*{Acknowledgements}
This work was partly supported by the National Key Research and Development Program of China(2020AAA0107600).  We would like to thank the anonymous reviewers for their constructive suggestions.

\bibliographystyle{ACM-Reference-Format}
\bibliography{sample-base}

\end{document}